%% file: main.tex
\documentclass[10pt,twocolumn,letterpaper]{article}

\usepackage{cvpr}
\usepackage{times}
\usepackage{epsfig}
\usepackage{graphicx}
\usepackage{amsmath}
\usepackage{amssymb}

\usepackage{times}
\usepackage{helvet}
\usepackage{courier}
\usepackage{graphicx}
\usepackage{bm}
\usepackage{color}
\usepackage{epstopdf}
\usepackage{caption}
\usepackage{subcaption}
\usepackage{enumitem}
\usepackage{calc}
\usepackage{multirow}
\usepackage{xspace}
\usepackage{booktabs}
\usepackage{mathrsfs}
\usepackage{array}
\usepackage{gensymb}
\usepackage{bbm}
\usepackage{dsfont}
\usepackage{makecell}

\newcommand{\thickhline}{%
    \noalign {\ifnum 0=`}\fi \hrule height 1pt
    \futurelet \reserved@a \@xhline
}

\usepackage[pagebackref=true,breaklinks=true,letterpaper=true,colorlinks,bookmarks=false]{hyperref}

\cvprfinalcopy 

\ifcvprfinal\fi
\begin{document}


\title{CVPR 2019 WAD Challenge on Trajectory Prediction and 3D Perception}



\author{Sibo Zhang\\
Baidu Research, Baidu Inc\\
{\tt\small \{sibozhang1\}@gmail.com}
\and
Yuexin Ma\\
Hong Kong Baptist University\\
{\tt\small \{yuexinma\}@comp.hkbu.edu.hk}
\and
Ruigang Yang\\
University of Kentucky\\
{\tt\small \{ryang2\}@uky.edu}
}


\maketitle

\begin{abstract}
   This paper reviews the CVPR 2019 challenge on Autonomous Driving. Baidu’s Robotics and Autonomous Driving Lab (RAL) providing 150 minutes labeled Trajectory and 3D Perception dataset including about 80k lidar point cloud and 1000km trajectories for urban traffic. The challenge has two tasks in (1) Trajectory Prediction and (2) 3D Lidar Object Detection. There are more than 200 teams submitted results on Leaderboard and more than 1000 participants attended the workshop. Workshop detail is here \footnote{\url{http://wad.ai/2019/challenge.html}}.
\end{abstract}

\input{introduction}
\input{dataset}

\input{challenge}

\input{method}

\input{conclusion}

	
	\section*{Acknowledgment}
This work is supported by Baidu Inc.. The authors gratefully all the participants of this challenge.

{\small
\bibliographystyle{ieee_fullname}

\bibliography{egbib}
}

\end{document}

%% file: introduction.tex
\section{Introduction}
\label{sec:intro}
The CVPR 2019 Workshop on Autonomous Driving — beyond single frame perception builds on the CVPR 2018 WAD with a focus on multi-frame perception, prediction, and planning for autonomous driving. It aims to get together researchers and engineers from academia and industries to discuss computer vision applications in autonomous driving. 

%% file: dataset.tex
\section{Dataset}
\label{sec:related}
Apolloscape Dataset~\cite{wang2019apolloscape} is a research-oriented project to foster autonomous driving in all aspects, from perception, navigation, prediction, to simulation. It hosts open access to semantically annotated (pixel-level) street view images and simulation tools that supports user-defined polices. Currently, it includes Trajectory Prediction~\cite{ma2019trafficpredict}, 3D Lidar Object Detection and 3D Lidar Object Tracking,  lanemark segmentation, online self-localization, 3D car instance understanding, Stereo, and Inpainting Dataset~\cite{liao2020dvi}. For each individual task, we set up an online evaluation platform and provide toolkit for users \footnote{\url{https://github.com/ApolloScapeAuto/dataset-api}}. 

For Trajectory Prediction and 3D Perception dataset, we use an acquisition car to collect traffic data, including camera-based images and LiDAR-based point clouds in the range of LiDAR. Our acquisition car runs in urban areas in rush hours. In the dataset, we provide camera images, 3D point cloud, and trajectories of traffic-agents in the LiDAR range. Our new dataset with 150min sequential data is a large-scale dataset for urban streets, which focuses on 3D perception, prediction, planning, and simulation tasks with heterogeneous traffic-agents. Details of dataset can be found at ~\cite{ma2019trafficpredict}. Trajectory dataset can be found at \footnote{\url{http://apolloscape.auto/trajectory.html}} and 3D Detection dataset at \footnote{\url{http://apolloscape.auto/tracking.html}}.

%% file: challenge.tex
\section{Challenge}
In this section, we introduce challenge details, evaluation metrics, and results. 

\subsection{Trajectory Prediction Challenge}

\textbf{Data Structure}
Our trajectory data is labeled at 2fps. For each row of the data file, it contains frame id, object id, the type of object, the position of object in the world coordinate system along x-axis, y-axis, and z-axis, the length of object, the width of object, the height of object, and the heading of object. The unit for the position and bounding box is meter. There are five different object types: 1 for small  vehicles, 2 for big  vehicles, 3 for pedestrian, 4 for motorcyclist and bicyclist, 5 for traffic cones, and 6 for others.

\textbf{Evaluation Metric}

During the evaluation, we treat the first two types, small vehicle and big vehicle, as one type (vehicle). In this challenge, the data from the first three seconds in each sequence is given as input data, the task is to predict trajectories of objects for the next three seconds. The objects used in evaluation are the objects that appear in the last frame of the first three seconds. The errors between predicted locations and the ground truth of these objects are then computed.

We adopt the metric similar to ~\cite{pellegrini2009you} to measure the performance of algorithms.

1. Average displacement error (ADE): The mean Euclidean distance over all the predicted positions and ground truth positions during the prediction time.

2. Final displacement error (FDE): The mean Euclidean distance between the final predicted positions and the corresponding ground truth locations.
Because the trajectories of vehicles, pedestrians, and bicyclist have different scales, we use the following weighted sum of ADE (WSADE) and weighted sum of FDE (WSFDE) as metrics.

\begin{equation}
    \label{eq:WSADE}
    WSADE = D_v\cdot ADE_v + D_p\cdot ADE_p + D_b\cdot ADE_b,
\end{equation}
\begin{equation}
    \label{eq:WSFDE}
    WSFDE = D_v\cdot FDE_v + D_p\cdot FDE_p + D_b\cdot FDE_b,
\end{equation}

where $D_v$, $D_p$, and $D_b$ are related to reciprocals of the average velocity of vehicles, pedestrian and bicyclist in the dataset. We adopt 0.20, 0.58, 0.22 respectively.

\subsection{3D Detection Challenge}

Our 3D Lidar object detection dataset consists of LiDAR scanned point clouds with high quality annotation. It is collected under various lighting conditions and traffic densities in Beijing, China. More specifically, it contains highly complicated traffic flows mixed with vehicles, cyclists, and pedestrians.

\textbf{Data Structure}
We labelled data for 3D Lidar object detection which file is a 1min sequence with 2fps. Each line in every file contains frame id, object id, object type, position x, position y, position z, object length, object width, object height, heading. The object types are defined the same as trajectory data. During the evaluation in this challenge, we treat the first two types, small vehicle and big vehicle, as one type (vehicle). Position is in the relative coordinate. The unit for the position and bounding box is meter. The heading value is the steering radian with respect to the direction of the object.

\textbf{Evaluation Metric}
We use similar metric defined in KITTI ~\cite{geiger2012we}. The goal in the 3D object detection task is to train object detectors for the classes 'vehicle', 'pedestrian', and 'bicyclist'. The object detectors must provide the 3D bounding box (3D dimensions and 3D position) and the detection score/confidence. We also note that not all objects in the point clouds have been labeled. We evaluate 3D object detection performance using mean average precision (mAP), based on IoU. Evaluation criterion similar to the 2D object detection benchmark (using 3D bounding box overlap). The final metric will be the mean of mAP of vehicles, pedestrian and bicyclist. We set IoU threshold for each type as 0.7 (Car), 0.5 (Pedestrian), 0.5(Cyclist).

%% file: method.tex
\section{Methods and Teams}
\subsection{Trajectory prediction}

\begin{figure*}[htbp]
	\centering
	\includegraphics[width=12cm]{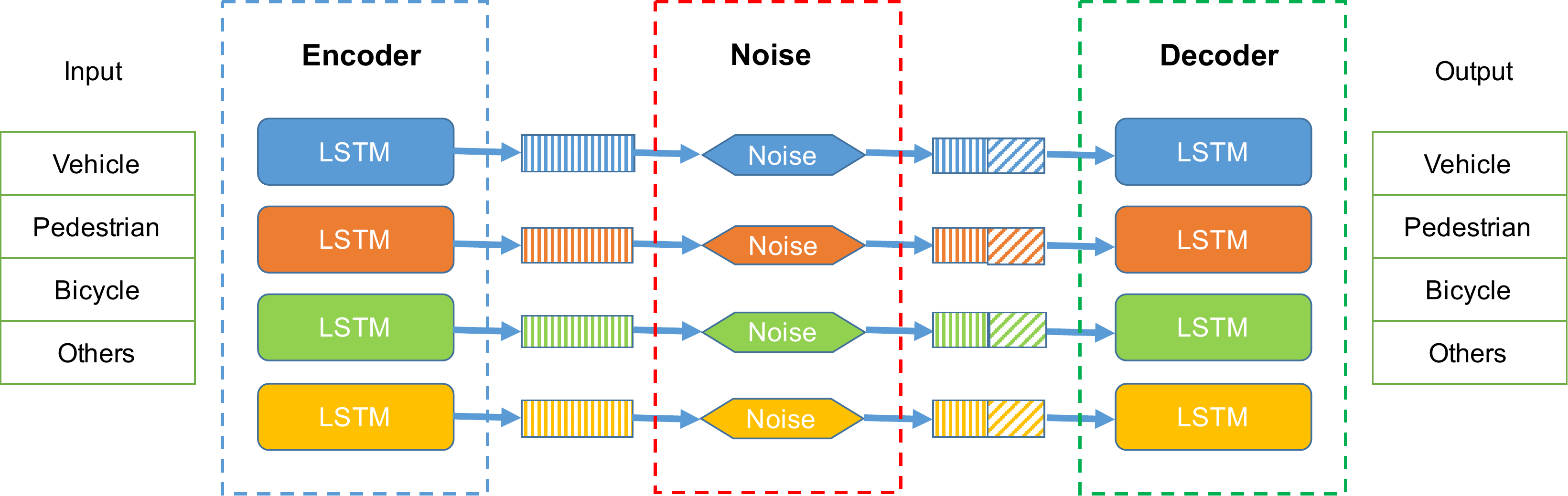}
	\caption{
		Multi-class LSTM Networks.
	}
	\label{fig:multi_class_lstm}
\end{figure*}

\textbf{MAD team (Xin Li, Yanliang Zhu, Deheng Qian and Dongchun Ren)} leveraged an LSTM-based encoder-decoder architecture for trajectory prediction task on urban street. More specifically, to make a more accurate forecasting, they applied four seq2seq sub-models to capture the motion properties of different traffic participants. As shown in Figure \ref{fig:multi_class_lstm}, they generated future trajectory for each agent through three stages, i.e., encoding, disturbing and decoding. Firstly, they used an encoder to embed the historical trajectory. Then, they introduced a 16-D random noise into the output tensor of the encoder to process the multi-modal distribution data. Finally, they outputted the prediction trajectory through a decoder which has the same structure as the encoder.

Besides, they also tried to the capture the collective influence among on-road agents using a global interaction technique as illustrated in \cite{zhu2019starnet}. Improving on the original method, they performed an interaction operation during the encoding and decoding stages at each time instant. The interaction module embedded the positions of all the agents and produced a global 128-D spatio-temporal representation by an LSTM unit. Then the calculated feature was sent to the encoders or decoders for the main prediction task. Each encoder or decoder, corresponding to a certain  individual, generated the private interaction within a local area by an attention operation using the above global feature and agent's location. According to their experimental results, interaction module could improve the prediction performance on the Apolloscape dataset.

\begin{figure*}[htbp]
	\centering
	\includegraphics[width = .9 \textwidth]{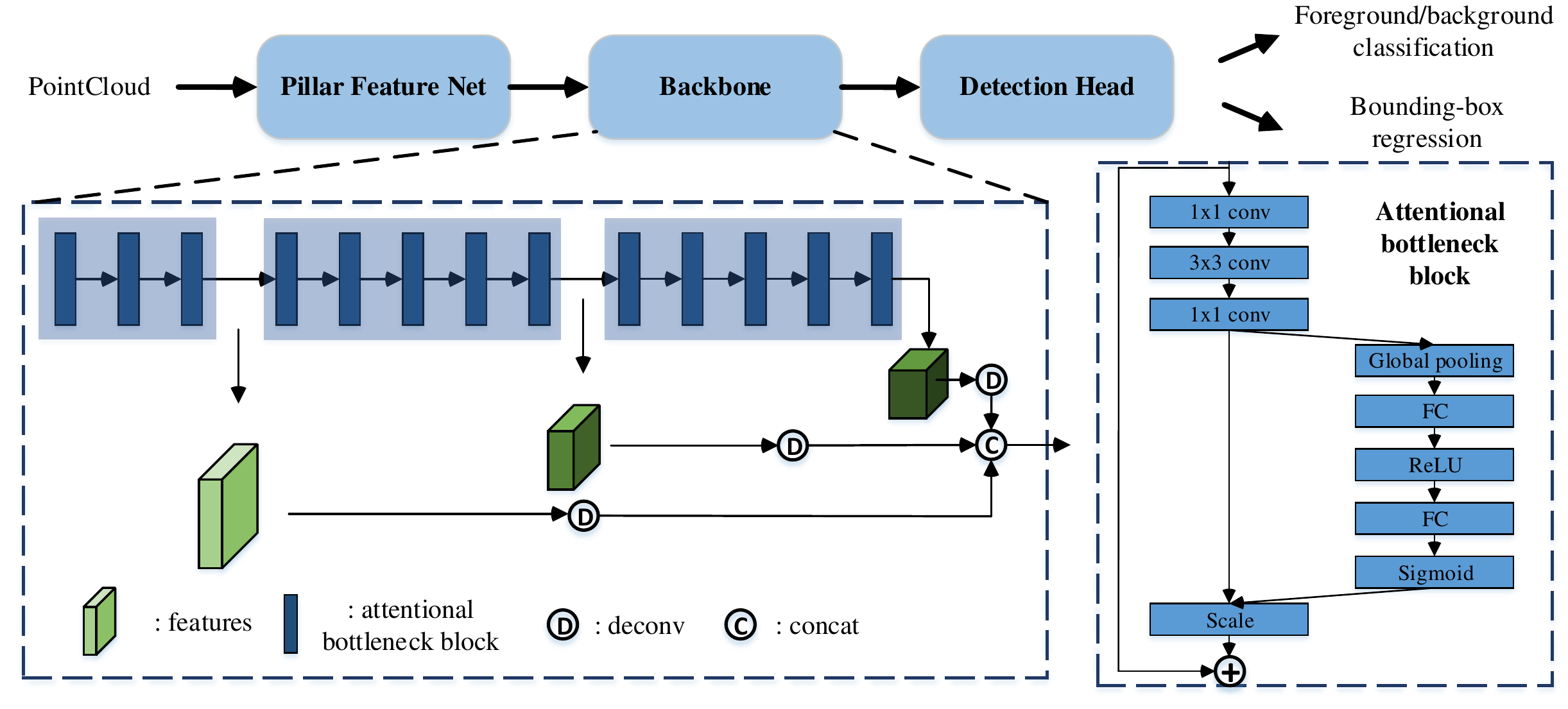}
	\caption{
		Illustration of the pipeline of the improved PointPillars method.
	}
	\label{fig:pipeline}
\end{figure*}

\subsection{3D Detection}

\begin{figure*}[htbp]
	\centering
	\includegraphics[width = .9 \textwidth]{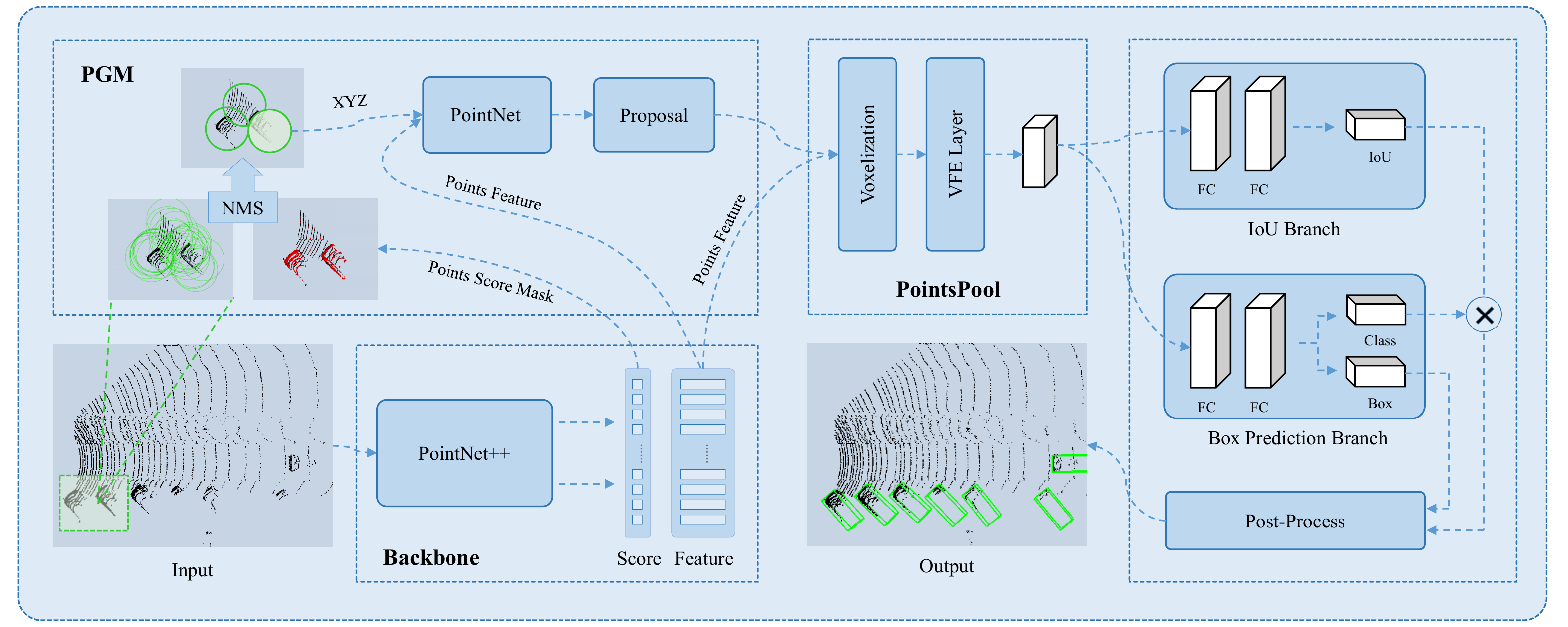}
	\caption{
		Illustration of the overall STD framework.
	}
	\label{fig:std}
\end{figure*}

\textbf{X-team (Zetong Yang, Yanan Sun, Shu Liu, Xiaoyong Shen, Jiaya Jia)} presented a novel method called sparse-to-dense 3D object detector (STD) \cite{YangSTD} . On the whole, STD is a double-stage point-based detection framework, which is illustrated in Figure \ref{fig:std}. The first stage of STD is a bottom-up proposal generation network. In this sub-network, they proposed to seed spherical anchors on each point to cover objects with different orientations. Since there is no need to consider objects with different orientations in generating anchors, this spherical design reduces the computation cost and brings less inference time. Then, interior points of these spherical anchors are gathered out to generate proposals for later refinement. In the second stage, a PointsPool layer is put forward to convert features of proposals from sparse point expressions to compact grid representations. These dense features are then passed through a prediction head which contains two extra fully-connected layers to generate final prediction results. They also provided a 3D intersection-over-union (IoU) branch in the prediction head to predict the 3D IoU between the final predictions and ground-truth bounding-boxes so as to retrieve predictions with more localization accuracy.

During training, they utilized 4 different data augmentations to prevent overfitting. First, similar to that of \cite{lang2019pointpillars}, they randomly add ground-truth bounding-boxes with their interior points from other scenes to current point cloud so as to simulate objects in various environments. Then, for each bounding box, they randomly rotate it with a uniform distribution $\Delta \theta_1 \in [- \pi / 4, + \pi / 4]$ and randomly add a translation ($\Delta x, \Delta y, \Delta z$). Third, each point cloud is randomly flipped along the $x$-axis with a probability of 0.5. Finally, randomly rotation and scaling are applied on each point cloud by a uniformly distributed random variable $\Delta \theta_2 \in [- \pi / 4, + \pi / 4]$ and a uniform distribution $[0.9, 1.1]$ respectively.
In the test phase, they first get prediction results on the original point cloud and the flipped point cloud on x-axis, and then ensemble these two results by Soft-NMS so as to generate final predictions.

\textbf{AK\_SG team's (Wenjing Zhang,  CherKeng Heng)} method is based on PointPillars~\cite{lang2019pointpillars}. The network setting is the mostly same to the original paper. They made modification to support more than one anchor for each class. The size of each class varies a lot(e.g. the width of car can be less than $0.3$ and larger than $11$), suggesting one anchor might not be enough. K-means is adopted to generate 5 anchors for each class. Another modification is that they disable the direction classification in the loss since the evaluation metric is based IOU, which is invariant to direction.  The point cloud range is set as $[-67.2, -50.4, -27, 67.2, 50.4, 5]$ for all classes.  The detail of other settings can be found in Table~\ref{tab:aksettings}.
\begin{table}[h]
\small
	\centering
	\resizebox{0.46\textwidth}{!}{
\begin{tabular}{c|c|c|c|c}
\hline
Class   & number of anchors & voxel size         & MNP & MNV   \\ \hline
Car     & 5                 & {[}0.28,0.28,32{]} & 50             & 20000                  \\ \hline
Bicyclist & 5                 & {[}0.14,0.14,32{]} & 20             & 80000                  \\ \hline
Pedestrian  & 5                 & {[}0.10,0.10,32{]} & 15             & 80000                  \\ \hline
\end{tabular}
}
\vspace*{3pt}
	\caption{The detailed setting of our method on each class. MNP denotes max num points and MNV represents max number of voxels. }
	\label{tab:aksettings}
\end{table}

For training data augmentation, they translate, scale the point cloud globally and rotate, translate each grounding truth. They disable global rotation of the point cloud since they surprisingly found it gives worse result. Detail parameters can be found in Table~\ref{tab:akaugmentation}.

\begin{table*}[t]
\small
	\centering
	\resizebox{0.95\textwidth}{!}{
\begin{tabular}{c|c|c|c|c}
\hline
Global Rotation & Global Translation & Global Scaling & Ground Truth Rotation & Ground Truth Tranlation    \\ \hline
{[}0,0{]}       & {[}0.2,0.2,0.2{]}  & {[}0.95,1.1{]} & {[}$-\frac{\pi}{20}$,$\frac{\pi}{20}${]}    & {[}0.25,0.25,0.25{]}      \\ \hline
\end{tabular}
}
\vspace*{3pt}
	\caption{The training data augmentation.}
	\label{tab:akaugmentation}
\end{table*}

\begin{figure}[t]
	\centering
	\includegraphics[width = .3 \textwidth]{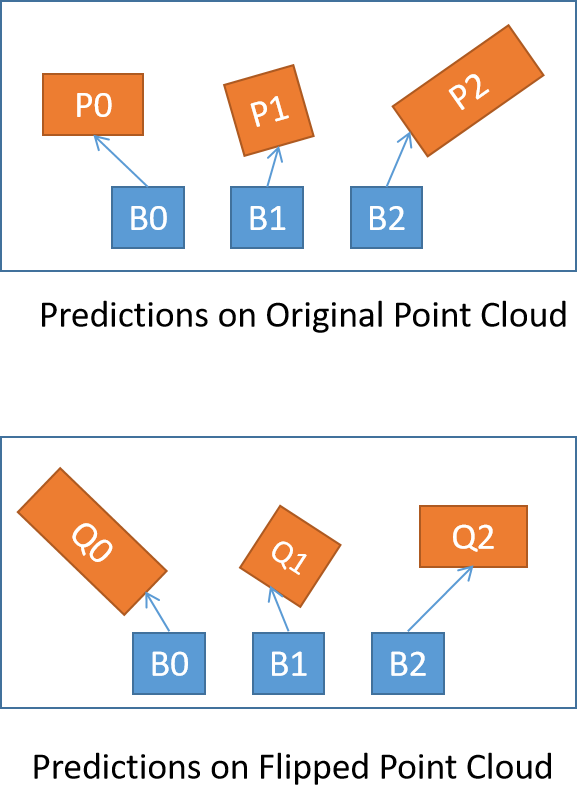}
	\caption{
		Illustration of Test Time augmentation. Two copies (original and flip-x) are used here. To do box fusion, we first flip back $Q0,Q1,Q2$.  Between point clouds, the regressed boxes have one to one correspondence. Here we have $(Q0,P2),(Q1,P1),(Q2,P0)$. Then we do box fusion for each pair. NMS is applied to the fused boxes.  
	}
	\label{fig:aktta}
\end{figure}

They adopted Test Time Augmentation to boost the performance. For each point cloud, they generate four copies: original, flip-x,flip-y, flip-xy. They feed each copy to the network and obtain regressed boxes.
They then flip the regressed boxes back. Since flip operation is adopted, the anchors between copies have one-to-one correspondence. For each anchor, they fuse the corresponding regressed boxes by averaging the box location, size and class probability. After that, NMS is used to remove redundant boxes. An illustration can be found in Fig.~\ref{fig:aktta}

\textbf{IAI-BIT team (Yuanpei Liu, Xingping Dong, Wenguan Wang and Jianbing Shen)} proposed a improved method based on PointPillars~\cite{lang2019pointpillars}. For the neural network structure, they introduced residual learning and channel attention~\cite{hu2018squeeze} to the baseline. As shown in the Figure~\ref{fig:pipeline}, our network consists of the original Pillar Feature Network, deeper 2D CNN backbone and a foreground/background classification\&regression detection head. Our backbone is much deeper than the original PointPillars and this brought big improvement on detection accuracy. For each class defined in the Apollo training dataset including: small vehicles, big vehicles, pedestrian, motorcyclist\&bicyclist, they trained a network to do foreground/background binary classification and thus there are a total of four networks. While generating final outputs, they combined all foreground predictions of these networks.
\begin{table}
	\small
	\centering
	\resizebox{0.47\textwidth}{!}{
		\begin{tabular}{c|c|c|c|c}
			\hline
			class&pointcloud range (m)&pillar size (m)&anchor size (m)&MSN\\
			\hline
			Vehicles&\makecell{x: ‐70.8$\sim$ 70\\y: ‐67.2$\sim$67.2\\z: ‐3$\sim$1}
			&\makecell{x: 0.16\\y: 0.16\\z: 3}&\makecell{x: 1.6\\y: 3.9\\z: 1.56}&15\\
			\hline
			Pedestrian&\makecell{x: ‐70.8$\sim$ 70\\y: ‐67.2$\sim$67.2\\z: ‐2.5$\sim$0.5}
			&\makecell{x: 0.2\\y: 0.2\\z: 3}&\makecell{x: 0.6\\y: 1.76\\z: 1.73}&15\\
			\hline
			Motor\&bicyclist&\makecell{x: ‐70.8$\sim$ 70\\y: ‐67.2$\sim$67.2\\z: ‐2.5$\sim$0.5}
			&\makecell{x: 0.2\\y: 0.2\\z: 3}&\makecell{x: 0.6\\y: 0.8\\z: 1.73}&15\\
			\hline
		\end{tabular}
	}
	\vspace*{3pt}
	\caption{The detailed setting of our method on each class. MSN denotes max sampling number}
	\label{tab:setting}
\end{table}
	
For the dataset pre-processing, they have tried the data augmentation methods used in KITTI~\cite{geiger2012we} dataset including: positive example sampling, global rotation, rotation for each object, random scale for each object. Different from KITTI~\cite{geiger2012we}, they adjusted the pre-processing method: not using the global rotation, narrowing down the range for scale and rotation, sampling more foreground point clouds from other samples to augment the positive examples. The detailed setting for each class can be seen in the Table~\ref{tab:setting}, which is suitable for Apollo dataset.

%% file: conclusion.tex
\section{Conclusion and Future Work}
In this paper, we reviews the CVPR 2019 challenge on Autonomous Driving. We analyze the  Competition underlying the 3D Detection and Trajectory prediction problems for autonomous driving-related research. We expect this paper to give cutting-edge insight in those areas. 

In the future, we will continue improve the ApolloScape open tool and dataset for autonomous driving. Besides, more workshops and challenges will be hosted to encourage exchanges of ideas and jointly advance the state of the art in autonomous driving research.

%% file: main.bbl
\begin{thebibliography}{1}\itemsep=-1pt

\bibitem{geiger2012we}
Andreas Geiger, Philip Lenz, and Raquel Urtasun.
\newblock Are we ready for autonomous driving? the kitti vision benchmark
  suite.
\newblock In {\em 2012 IEEE Conference on Computer Vision and Pattern
  Recognition}, pages 3354--3361. IEEE, 2012.

\bibitem{hu2018squeeze}
Jie Hu, Li Shen, and Gang Sun.
\newblock Squeeze-and-excitation networks.
\newblock In {\em Proceedings of the IEEE conference on computer vision and
  pattern recognition}, pages 7132--7141, 2018.

\bibitem{lang2019pointpillars}
Alex~H Lang, Sourabh Vora, Holger Caesar, Lubing Zhou, Jiong Yang, and Oscar
  Beijbom.
\newblock Pointpillars: Fast encoders for object detection from point clouds.
\newblock In {\em Proceedings of the IEEE Conference on Computer Vision and
  Pattern Recognition}, pages 12697--12705, 2019.

\bibitem{liao2020dvi}
Miao Liao, Feixiang Lu, Dingfu Zhou, Sibo Zhang, Wei Li, and Ruigang Yang.
\newblock Dvi: Depth guided video inpainting for autonomous driving.
\newblock {\em arXiv preprint arXiv:2007.08854}, 2020.

\bibitem{ma2019trafficpredict}
Yuexin Ma, Xinge Zhu, Sibo Zhang, Ruigang Yang, Wenping Wang, and Dinesh
  Manocha.
\newblock Trafficpredict: Trajectory prediction for heterogeneous
  traffic-agents.
\newblock In {\em Proceedings of the AAAI Conference on Artificial
  Intelligence}, volume~33, pages 6120--6127, 2019.

\bibitem{pellegrini2009you}
Stefano Pellegrini, Andreas Ess, Konrad Schindler, and Luc Van~Gool.
\newblock You'll never walk alone: Modeling social behavior for multi-target
  tracking.
\newblock In {\em 2009 IEEE 12th International Conference on Computer Vision},
  pages 261--268. IEEE, 2009.

\bibitem{wang2019apolloscape}
Peng Wang, Xinyu Huang, Xinjing Cheng, Dingfu Zhou, Qichuan Geng, and Ruigang
  Yang.
\newblock The apolloscape open dataset for autonomous driving and its
  application.
\newblock {\em IEEE transactions on pattern analysis and machine intelligence},
  2019.

\bibitem{YangSTD}
Zetong Yang, Yanan Sun, Shu Liu, Xiaoyong Shen, and Jiaya Jia.
\newblock {STD:} sparse-to-dense 3d object detector for point cloud.
\newblock In {\em 2019 {IEEE/CVF} International Conference on Computer Vision,
  {ICCV} 2019, Seoul, Korea (South), October 27 - November 2, 2019}, 2019.

\bibitem{zhu2019starnet}
Yanliang Zhu, Deheng Qian, Dongchun Ren, and Huaxia Xia.
\newblock Starnet: Pedestrian trajectory prediction using deep neural network
  in star topology.
\newblock {\em IROS}, 2019.

\end{thebibliography}
